\documentclass{article}

\usepackage[preprint, nonatbib]{neurips_2020}

\usepackage[utf8]{inputenc} 
\usepackage[T1]{fontenc}    
\usepackage{hyperref}       
\usepackage{url}            
\usepackage{booktabs}       
\usepackage{amsfonts}       
\usepackage{nicefrac}       
\usepackage{microtype}      

\usepackage{graphicx} 
\usepackage{makecell} 
\usepackage{adjustbox} 
\usepackage{subcaption} 

\usepackage{pifont}
\newcommand{\cmark}{\ding{51}}%
\newcommand{\xmark}{\ding{55}}%

\usepackage{colortbl}

\usepackage{pgfplots}
\usepgfplotslibrary{fillbetween}

\usepackage{wrapfig}

\newcommand{\etal}{\textit{et al}.}

\title{Surprisal-Triggered Conditional Computation with Neural Networks}

\author{%
Loren Lugosch$^{1,2}$, Derek Nowrouzezahrai$^{1,2}$, Brett H. Meyer$^1$ \\
  $^1$McGill University, $^2$Mila\\
  \texttt{lugoschl@mila.quebec}, \texttt{derek@cim.mcgill.ca}, \texttt{brett.meyer@mcgill.ca}  \\
}

\begin{document}

\maketitle

\begin{abstract} 
Autoregressive neural network models have been used successfully for sequence generation, feature extraction, and hypothesis scoring. This paper presents yet another use for these models: allocating more computation to more difficult inputs. In our model, an autoregressive model is used both to extract features and to predict observations in a stream of input observations. The surprisal of the input, measured as the negative log-likelihood of the current observation according to the autoregressive model, is used as a measure of input difficulty. This in turn determines whether a small, fast network, or a big, slow network, is used. Experiments on two speech recognition tasks show that our model can match the performance of a baseline in which the big network is always used with 15\% fewer FLOPs.
\end{abstract}

\section{Introduction}
\label{intro}

In ``Thinking, Fast and Slow'', Daniel Kahneman hypothesizes that human cognition operates in one of two modes: ``System 1'' cognition, which is fast, automatic, and effortless, and ``System 2'' cognition, which is slow, deliberate, and effortful \cite{kahneman2011thinking, stanovich2000individual}. What determines whether System 1 or System 2 is active at a given time is roughly the current level of cognitive ease: most of the time System 1 dominates, and only when something breaks down and the environment becomes difficult to predict or control does System 2 activate. An example of experimental support for this hypothesis, or at least for the weaker hypothesis that environmental surprisal controls cognitive effort in some way, can be found in studies of reading time: words that are surprising (in the sense that a statistical language model assigns lower probability to them), as well as the words that follow, require more time for human subjects to read, suggesting that more effort is being used \cite{levy2008expectation, levy2011integrating, monsalve2012lexical}. 

In contrast to human cognition, the deep neural networks used in artificial intelligence typically do not perform any less computation for any input: the model always multiplies the input by the same sequence of weight matrices to compute an output, no matter how difficult or easy the input may be. 
This seems like a waste of energy. As neural networks have gotten bigger \cite{amodei31ai, sanh2019distilbert} and more expensive to run \cite{strubell2019energy, sharir2020cost}, it has become more pressing to find ways to address this waste.
The question this paper asks is: can we emulate human cognition to improve the computational efficiency of neural networks?

\begin{figure}[t!]
    \centering
    \includegraphics[scale=0.8]{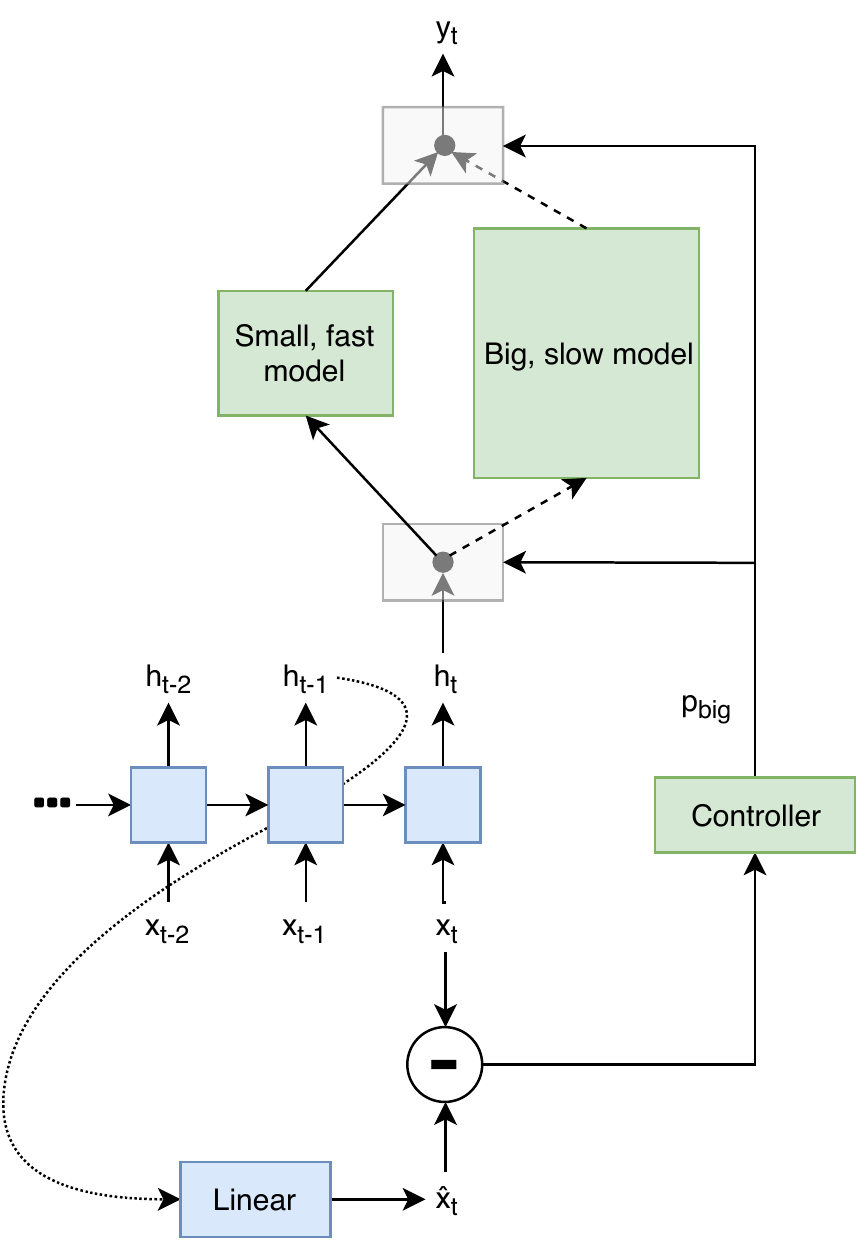}
    \caption{Architecture of the proposed system. Here the autoregressive component of the model (blue) is depicted as an RNN.}
    \label{fig:net}
\end{figure}

To try to answer this question, we present a simple model of conditional computation for neural networks that mirrors the System 1/System 2 division of labor. The model is depicted in Figure \ref{fig:net}. An autoregressive neural network model is used to process a stream of input observations, both to extract features and to predict the next observation. A small, fast network, loosely analogous to System 1, runs most of the time, minimizing the amount of computation that must be performed on average, and a big, slow network, loosely\footnote{Here we are not considering more ``logical'' operations, like sequential reasoning and symbol manipulation; rather, we are focusing on the aspects of this model relevant to the control of computational resources.} analogous to System 2, runs only when the autoregressive component is unable to accurately predict the current input.

In addition to resembling certain theories about human cognition, our model can be thought of as taking advantage of the ``low-density separation'' assumption of semi-supervised learning \cite[p.7]{chapelle2009semi}: namely, that the optimal decision boundary for a classifier lies in a low-density region of the input space. The autoregressive model in our setup explicitly detects when the input is in such a region. Under the  low-density separation assumption, the less surprising the input is, the farther away the input is from the complex decision boundary defined by the big network. Hence, it may be expected that less error will be incurred when approximating that decision boundary using the small network---though for now, we have no formal proof of how good this approximation might be.

In our experiments, we observe an improved tradeoff between computation and accuracy using our model: it can perform as well as or even better than models that always use the big network at lower cost. The improvement is consistent over a variety of hyperparameter settings for two datasets, which gives strong evidence that surprisal is a useful inductive bias for conditional computation. Another interesting contribution this paper makes is that while other work has used autoregressive models \textit{either} as pre-trained feature extractors \textit{or} for predicting future observations, our work seems to be the first to show that it is possible and worthwhile to use them for \textit{both} functions at the same time. 

\section{Related Work}

\paragraph{Conditional computation}
In conditional computation, only a fraction of a model's parameters are used to process any given input. Some machine learning algorithms, like decision trees, natively support conditional computation, but neural networks do not. In the last few years, researchers have begun thinking more about how to incorporate conditional computation into neural networks \cite{bengio2013estimating, davis2013low}.

Perhaps the two most commonly used approaches to conditional computation with neural networks are the early exiting and score margin techniques. 
In early exiting, a classifier is attached to an intermediate layer of a network and trained to determine whether stopping at that layer will result in a misclassification; if a misclassification is not predicted, subsequent layers are not computed \cite{teerapittayanon2016branchynet, tan2016towards, Xin2020deebert, stamoulis2018designing, scardapane2020add}.
In the score margin approach, a small model is used to compute scores for each class in a classification problem; if the margin between the largest score and the second largest score is below a certain threshold, then a bigger model is used \cite{park2015big, gruenstein2017cascade, tann2018flexible, pagliari2018dynamic, venkataramani2015scalable}.

In other approaches, the model itself learns when to use its various components. This often takes the form of a mixture-of-experts \cite{jacobs1991adaptive, masoudnia2014mixture}, in which a learned controller is used to select the experts relevant for a given input \cite{bengio2015conditional, leonard2015distributed, yu2017learning, odena2017changing, liu2018dynamic, dean2019deep}, possibly hierarchically \cite{eigen2013learning, cho2014exponentially, rosenbaum2017routing, tanno2018adaptive}. The binary decision of selecting or not selecting an expert is not differentiable, so it is not possible to perform standard backpropagation in a mixture-of-experts. Instead, these approaches treat the expert selection as a policy, and use reinforcement learning to train the policy. To avoid the difficulties of reinforcement learning, often a soft approximation to the hard selection decision is used during training \cite{bengio2013estimating, graves2016adaptive, figurnov2017spatially, shazeer2017outrageously, campos2017skip, jernite2016variable, bolukbasi2017adaptive, dehghani2018universal, bapna2020controlling}.

The model we propose strikes a balance between the more ``innate'' and the more ``learned'' approaches to conditional computation. It is hard-wired to use surprisal to determine when to use the bigger network, but this measure of surprisal is learned in an unsupervised way. 
Unlike more innate approaches, our model makes few assumptions about the nature of the problem or domain: only that it is sensible to express the input as a sequence of observations, which is true of many data modalities, like audio, text, and video. 
Unlike more learned approaches, our model is more suitable for scenarios where there is very limited labeled training data (as long as there is sufficient unlabeled data available to train the autoregressive model).

\paragraph{Surprisal-based models and autoregressive models}
Surprisal is a useful notion for many tasks, such as anomaly detection \cite{NIPS2019_9611} and comparing the difficulty of modeling different languages \cite{mielke2019kind}. Using surprisal as an input feature in a neural network model has been explored in the past: for instance, in \cite{he2019pun}, He He \etal~generate puns using a model with surprisal-based features. \cite{rocki2016surprisal} uses prediction error as an input to the model, though not for the purposes of conditional computation. In \cite{rocki2016zoneout, alpay2019preserving}, surprisal is used to determine whether to apply zoneout to units in LSTMs. 

Autoregressive neural network models have not only been used to predict future observations or as generative models; they have also been used more recently with great success as unsupervised pre-trained feature extractors \cite{dai2015semi, lotter2016deep, radford2017learning, howard2018universal, ha2018world, chung2019unsupervised, yang2019xlnet}. This seems to work well because accurately predicting the future necessitates extracting informative features from the past \cite{srivastava2015unsupervised, radford2019language}. Hence, we use the autoregressive model in our setup not only for measuring surprisal, but also to preprocess the input, which amortizes the additional cost incurred by running the autoregressive model.

\paragraph{Other related ideas}
Our model is similar to certain techniques in data compression and signal processing. In arithmetic coding, less effort is allocated to less surprising inputs, in the sense that shorter bitstrings (less bandwidth) are assigned to more predictable symbols \cite{rissanen1976generalized, witten1987arithmetic}. In linear predictive coding, a linear filter is used to predict the next sample of the input signal from previous samples, and the filter coefficients and error signal are transmitted instead \cite{frantz1982design, o1988linear, oord2018representation}.
According to the predictive coding hypothesis in neuroscience, human brains communicate information in a similar way: not as the raw signals themselves but rather in the form of prediction error \cite{huang2011predictive, clark2013whatever}.

The Neural Sequence Chunker model proposed in \cite{schmidhuber1991neural} is very similar to our model, in that the prediction error of an autoregressive model is used to control a subsequent model; but in that work, the subsequent model simply does not process predictable inputs at all, so as to reduce the input sequence length, whereas here we assume that every input must result in a corresponding output, which is often the case in sequence modeling. 
Another idea related to our model is the Expert Gate \cite{aljundi2017expert}, which uses the reconstruction error of a set of autoencoders to select an expert in a mixture-of-experts to use. 
In \cite{sheikhnezhad2019novel}, a reinforcement learning agent uses both a habitual controller and a planning-based controller and arbitrates between them using state prediction error and reward prediction error. Similarly, the Variational Bandwidth Bottleneck of \cite{goyalvariational} uses a notion of channel capacity to determine whether to run an expensive model-based planner.
Our model is somewhat simpler and more broadly applicable than these approaches; it can be used outside of reinforcement learning and makes no constraints on the exact nature of the big and small networks.

\section{Model Architecture}

Here we describe how our model works in more detail. Overall, the input to the model is a sequence of observations $(x_{1},x_2,x_{3},\dots)$, and for each timestep $t$, the model produces an output $y_t$. 

\subsection{Autoregressive model}

An autoregressive model expresses the joint distribution of a sequence $p(x_1, x_2, x_3, \dots)$ as the product of the conditional distributions of the elements of the sequence given the previous elements:

\begin{equation}
    p(x_1, x_2, x_3, \dots) = \prod_t p(x_t | x_{t-1}, x_{t-2}, \dots ),
\end{equation}
where $p(x_t | x_{t-1}, x_{t-2}, \dots )$ can be estimated using a neural network  \cite{elman1990finding, bengio2003neural, larochelle2011neural}. 

The surprisal of an observation is defined as the negative log-likelihood of that observation under the distribution defined by the autoregressive model. For real-valued inputs, a reasonable choice for the distribution is an isotropic Gaussian with variance 1, in which case surprisal is equivalent (minus a constant term) to the squared error between the model's prediction $\hat{x}_{t}$ and the actual observation $x_{t}$:

\begin{equation}
    \textrm{Surprisal}(x_{t}) = -\log p(x_{t} | x_{t-1}, x_{t-2}, \dots ) \approx \frac{1}{2} ||x_{t} - \hat{x}_{t}||_2^2.
\end{equation}

We use a neural network encoder to compute a feature vector $h_t = f(x_t, x_{t-1}, \dots)$ and a linear model to compute the prediction $\hat{x}_{t}$ of the observation $x_{t}$ given the feature vector from the previous timestep $h_{t-1}$. In our experiments, we use RNNs to implement the encoder, but other causal neural network layers, like causal convolutions \cite{gehring2017convolutional} and masked self-attention \cite{vaswani2017attention}, could be used as well.

Because the autoregressive model is unsupervised, it could be trained either using the input data for the target task or on a larger source of unlabeled data in the same domain. For example, in our experiments, we train conditional computation models on the small TIMIT dataset, but we train the autoregressive part on the much larger LibriSpeech dataset. 

It is natural to ask whether it is worthwhile to backpropagate through the entire model, including the autoregressive model, when training on the downstream task \cite{peters2019tune}. In the experiments for this paper, however, we simply keep the weights of the autoregressive model frozen. The reason is that if its weights are trained along with the rest of the model, it may not remain autoregressive, in which case it will not accurately compute surprisal. It should be possible to jointly train the entire model for better performance by adding a term to the final loss function that encourages the autoregressive model to remain autoregressive, but then this term would require its own regularization strength, which would mean another hyperparameter to be tuned, adding further complexity and variability to our experiments.

\subsection{Controller}
The controller uses the surprisal of the current observation $x_{t}$ to compute the probability of sampling the big network $p_{big}$:
\begin{equation}
    p_{big} = \textrm{sigmoid}(w \cdot \textrm{Surprisal}(x_{t}) + b),
\end{equation}
where $w$ and $b$ are scalars. 

We may want the distribution of $p_{big}$ to have a certain mean (to achieve a certain computational budget) and variance (to ensure that both networks have the chance to be sampled for any given input, instead of only greedily sampling one or the other). We can train the controller so that $p_{big}$ has a specified mean $\mu$ and variance $\sigma^2$ by minimizing the following loss function with respect to the controller parameters $w$ and $b$:

\begin{equation}
    \mathcal{L}_{controller} = \frac{1}{2}(\hat{\mu} - \mu)^2 + \frac{1}{2}(\hat{\sigma}^2 - \sigma^2)^2,
\end{equation}

where $\hat{\mu}$ and $\hat{\sigma}^2$ are the sample mean and variance of $p_{big}$. Alternately, one could specify a target budget in terms of FLOPs, and set $w$ and $b$ so that the resulting expected amount of computation is equal to this budget. Like the autoregressive encoder, training the controller can be done either using the target dataset/environment or using a separate stream of unlabelled data.

\subsection{Big and small networks}
We use simple fully-connected neural networks to implement the big and small networks. An interesting aspect of the model is that if gradients are not backpropagated into the autoregressive model, non-gradient-based learners like decision trees could easily be used here, similar to the way that evolutionary algorithms are used in conjunction with neural world models in \cite{ha2018world}.

One caveat for the big and small networks is that it may not be straightforward to implement them using stateful models. For example, if we were to use RNNs as the big and small networks, where the state vector for the small RNN is not the same size as the state vector for the big RNN, it would not be possible to switch between these networks without introducing some additional machinery. 
A workaround that makes it easier to use stateful models is described in the next subsection. In the future, it could be interesting to find ways to overcome this limitation, possibly using models like Neural ODEs \cite{chen2018neural} that can maintain state across arbitrary timespans.

\subsection{Pre-net and post-net}
It is optionally possible to sandwich the conditional part of the model between a non-conditional ``pre-net'' and ``post-net'', as was done in the Sparsely Gated Mixture-of-Experts of \cite{shazeer2017outrageously}. In this case, the output of the autoregressive model $h_t$ is instead fed to the pre-net, and the output $y_t$ is taken from the post-net. This could be used to more easily add state, for example, since the pre-net and post-net can be implemented using RNNs.

\section{Experiments}
There are a number of questions one might ask about our model. How much computation does it save? Is surprisal actually a good heuristic for effort allocation? In other words, do we get better results using surprisal-triggered sampling than if we were to learn a controller or to just sample the big or small networks at random? Is our decision to use the features computed by the autoregressive model instead of the original inputs justified? How robust are the results to the choice of hyperparameters? We ran experiments to answer each of these questions.

\subsection{Datasets}
We use two small speech recognition datasets for our experiments\footnote{Our PyTorch \cite{paszke2019pytorch} experiment code can be found online at \url{https://github.com/lorenlugosch/conditional-computation-using-surprisal}.}: TIMIT and Mini-LibriSpeech. 
TIMIT \cite{garofolo1993darpa} is a 3-hour dataset with hand-aligned phoneme labels. Mini-Librispeech\footnote{\url{https://www.openslr.org/31/}} is a 5-hour subset of the 960-hour LibriSpeech dataset \cite{panayotov2015librispeech} with transcripts but no phoneme labels; we used the Montreal Forced Aligner \cite{mcauliffe2017montreal} to obtain label sequences from the transcripts.

\subsection{Setup}
The small network in these experiments is a fully connected leaky ReLU layer with 512 hidden units. The big network is a fully connected layer with 2048 hidden units followed by another fully connected layer with 512 hidden units. The pre-net is a bidirectional GRU \cite{chung2014empirical} with 256 hidden units in each direction and 50\% dropout \cite{zaremba2014recurrent}. The post-net uses the same architecture as the pre-net, followed by a fully connected layer with ($n$ + 1) outputs, where $n$ is the number of phonemes (39 for TIMIT, 41 for Mini-LibriSpeech) and 1 is for the CTC ``blank'' symbol.

The autoregressive model has two unidirectional GRU layers with 512 hidden units and 50\% dropout, each followed by a fully connected layer with 512 hidden units. The inputs to the autoregressive model are sequences of 80-dimensional filterbank frames extracted using a 25 ms Hamming window every 10 ms from the 16,000 Hz audio signal. We skip every second frame \cite{sak2015fast}, since this made our experiments much faster to run at a small cost in accuracy. We train the autoregressive model with maximum likelihood estimation on the full 960 hours of LibriSpeech.

The controller is trained for TIMIT to have mean 0.5 and variance 0.04 for $p_{big}$, which is the mean and variance that result from a standard normal distribution for the input to the sigmoid. 
Having this distribution ensures that $p_{big}$ does not saturate at 0 or 1 and the range in between 0 and 1 is covered, so a given input observation is not always presented to only the big network or only the small network. For Mini-LibriSpeech, we instead train the controller to have mean 0.65 for $p_{big}$ to bias it more towards using the big network, since its validation set has much lower surprisal. We use one pass of SGD through the training set of the task of interest to train the controller. 

The number of parameters (which for this architecture happens to translate exactly to the number of FLOPs\footnote{A FLOP can be defined in different ways. We use the convention that one multiply-accumulate = one FLOP. (FLOP count does not always precisely correspond to an actual useful metric, like latency or power consumption, but we leave more realistic evaluations like these for the future.)} performed by the network for each input timestep) for each part of the model is shown in Table \ref{parameter-count}. Each model is trained using CTC \cite{graves2006connectionist} for 50 epochs. The validation performance is measured at the end of every epoch or every 5 epochs (we do this for Mini-LibriSpeech to speed up our experiments, since decoding its validation set is much slower than training for one epoch). The model checkpoint with the best validation phoneme error rate over the course of training is used for the test set \cite{hannun_2017}. The test set is decoded using a beam search of width 10. Each model is trained with 5 random seeds, and we report the mean and standard deviation of results over the 5 trials.

\begin{table}
  \caption{Parameter counts for model used in main experiments.}
  \label{parameter-count}
  \centering
  \begin{tabular}{lr}
    \toprule
    Component     & Number of parameters \\
    \midrule
    Autoregressive model &  3.05M  \\ 
    Pre-net     & 1.18M \\ 
    Small network    & 0.26M \\ 
    Big network  & 2.10M \\ 
    Post-net     & 1.20M\\ 
    Controller & 2\\
    \bottomrule
  \end{tabular}
\end{table}

\subsection{Comparison with existing conditional models}
\label{sec:compare}
We first compare our model with existing techniques for conditional computation. Early exiting is not applicable here because our experiments use CTC models, not classifiers, so it is not possible to implement the misclassification predictor; likewise, the score margin technique is not applicable because there are more intermediate stages between the big and small networks and the softmax output. We therefore compare with a simple baseline model that is identical in every respect, except that a learned controller is used instead of using surprisal. The learned controller is a feedforward gating network which takes as input $h_t$ and outputs $s_t$, a binary decision to select the big network or small network. The network has a single hidden layer with 80 hidden units so that this controller is roughly the same size as the linear model used in the autoregressive model to estimate $\hat{x}_t$ (which is 80-dimensional) from $h_{t-1}$. We use the straight-through estimator \cite{bengio2013estimating}: in other words, we treat the threshold function used to compute $s_t$ as the identity function during backpropagation so that the controller receives a non-zero gradient. The overall loss function used for training this model is:
\begin{equation}
    \mathcal{L} = \mathcal{L}_{CTC} + \lambda \cdot \sum_t (s_t - 0.5)^2,
\end{equation}
where the second term encourages the model to use the big network roughly half of the time, and $\lambda = 0.001$ (chosen using a grid search in \{0.1, 0.01, 0.001, 0.0001, 0.00001\}). 

The results of the experiment---the phoneme error rate (PER) and average FLOPs per input timestep for the test set of TIMIT and Mini-LibriSpeech---are shown in Table \ref{comparison}. The learned controller and the surprisal-based controller have similar performance; however, the surprisal-based model has lower variance in PER and a lower cost in FLOPs for both datasets. The performance of the learned controller model also seems to be sensitive to the more opaque hyperparameter $\lambda$: when $\lambda$ is set to 0.0001 instead of 0.001, its PER and FLOPs for TIMIT increase to 23.97\% and 6.99M, respectively (see Appendix D). In contrast, the $\mu$ and $\sigma^2$ target hyperparameters for the surprisal-based controller are easy to interpret and do not need an expensive grid search to be set. This is an important consideration because when training extremely large models---where conditional computation may prove especially useful---there is often only enough budget for a small number of training runs \cite{brown2020language}.

\begin{table}[]
\centering
\caption{Results for TIMIT and Mini-LibriSpeech. (For all tables, rows in grey are Pareto-optimal.)}
\label{comparison}
\begin{tabular}{ccccc} 
\toprule
& \multicolumn{2}{c}{TIMIT}  & \multicolumn{2}{c}{Mini-LibriSpeech}                  \\
    \cmidrule(r){2-3} \cmidrule(r){4-5}
\thead{Model} & \thead{PER} & \thead{Avg. FLOPs\\per input} & \thead{PER} & \thead{Avg. FLOPs\\per input} \\
\midrule
Random controller & 20.52\% $\pm$ 0.28\% &  6.63M & 26.16\% $\pm$ 0.35\%  &  6.62M \\
Learned controller & \cellcolor[gray]{.75}\textbf{19.91\%} $\pm$ 0.58\% & \cellcolor[gray]{.75}6.63M  & 25.93\% $\pm$ 0.32\%  & 6.50M \\ 
Surprisal-based controller & \cellcolor[gray]{.75}20.00\% $\pm$ 0.15\% & \cellcolor[gray]{.75}\textbf{6.41M} & \cellcolor[gray]{.75}25.80\% $\pm$ 0.10\%  & \cellcolor[gray]{.75}\textbf{6.43M} \\
\midrule
Big network only &  20.09\% $\pm$ 0.24\%   & 7.54M & \cellcolor[gray]{.75}\textbf{25.69\%} $\pm$ 0.30\%  & \cellcolor[gray]{.75}7.54M  \\ 
\bottomrule
\end{tabular}
\end{table}

\subsection{Ablation study}
Our model essentially makes three independent design choices: it uses features computed by the autoregressive model instead of the original inputs, it samples the big network according to surprisal (as opposed to just randomly with probability 0.5) during training, and it samples the big network according to surprisal during testing. We trained models where each of these choices is ablated and report the results, as well as the baseline results when just the small network or just the big network is used. The results\footnote{Note that the FLOP counts are lower for the models that do not use autoregressive features because the input-to-hidden weight matrices in the pre-net have input dimension 80 (the dimension of the filterbank features) instead of 512 (the dimension of the autoregressive features). Also, the FLOP counts for the first and third rows are lower by 3.05M because for them the autoregressive model does not need to be run at all during test time.} of the ablation study with TIMIT are shown in Table \ref{ablation-study}.

\begin{table}[]
\centering
\caption{Results of ablation study for TIMIT.}
\label{ablation-study}
\begin{tabular}{ccccc} 
\toprule
\thead{Autoregressive\\ features?} & \thead{Surprisal-based\\ during training?} & \thead{Surprisal-based\\ during testing?} & \thead{PER} & \thead{Avg. FLOPs\\per input}\\
\midrule
\xmark  & \xmark & \xmark    &  22.81\% $\pm$ 0.27\%   &   \textbf{2.91M}    \\
\xmark  & \xmark & \cmark   &  22.49\% $\pm$ 0.28\%  &  5.75M \\
\rowcolor[gray]{.75}\xmark  & \cmark  & \xmark   &  22.73\% $\pm$ 0.44\%  & \textbf{2.91M} \\
\xmark & \cmark  & \cmark    &  22.58\% $\pm$ 0.05\%  &  5.75M \\ 
\cmark & \xmark & \xmark   & 20.52\% $\pm$ 0.28\% &  6.63M  \\ 
\cmark  & \xmark  & \cmark  &  20.52\% $\pm$ 0.17\% & 6.41M  \\
\cmark & \cmark & \xmark  & 20.28\% $\pm$ 0.17\% & 6.63M \\
\rowcolor[gray]{.75}\cmark & \cmark  & \cmark   & \textbf{20.00\%} $\pm$ 0.15\% &  6.41M \\ 
\midrule
\rowcolor[gray]{.75}\multicolumn{3}{c}{Small network only}  & 20.61\% $\pm$ 0.24\% &  5.70M  \\ 
\multicolumn{3}{c}{Big network only} &  20.09\% $\pm$ 0.24\%   & 7.54M \\ 
\bottomrule
\end{tabular}
\end{table}

First, we find that across all experiments, using the features computed by the autoregressive model results in significantly lower PER than using the original input features. This is perhaps not surprising, but it does independently confirm the efficacy of filterbank-based autoregressive models as pre-trained feature extractors for speech recently proposed in \cite{chung2019unsupervised}. For this reason, we use the autoregressive features in subsequent experiments and when using the big network only or the small network only (the last two rows of Tables \ref{ablation-study} and \ref{ablation-study-librispeech}).

Next, we find that surprisal-based sampling yields Pareto-optimal results, with performance closer to or even slightly outperforming the models that only use the big network compared to the models that do not use surprisal. Similar results are obtained for Mini-LibriSpeech (Table \ref{ablation-study-librispeech}) and when using other neural network architectures (Appendix A). Also, if only the models not using autoregressive features are considered (the first four rows of Table \ref{ablation-study}), the model with surprisal-based sampling during both training and testing is still Pareto-optimal. 

\begin{table}[]
\centering
\caption{Results of ablation study for Mini-LibriSpeech.}
\label{ablation-study-librispeech}
\begin{tabular}{cccc} 
\toprule
\thead{Surprisal-based\\ during training?} & \thead{Surprisal-based\\ during testing?} & \thead{PER} & \thead{Avg. FLOPs\\per input}\\
\midrule
\xmark & \xmark     & 26.16\% $\pm$ 0.35\%  &  6.62M     \\ 
\xmark  & \cmark  &  26.04\% $\pm$ 0.26\%    &     6.43M  \\ 
\cmark & \xmark  &  26.44\% $\pm$ 0.06\%  &  6.62M     \\ 
\rowcolor[gray]{.75}\cmark & \cmark  & 25.80\% $\pm$ 0.10\%  & 6.43M   \\ 
\midrule
\rowcolor[gray]{.75}\multicolumn{2}{c}{Small network only}   & 26.31\% $\pm$ 0.33\% &  \textbf{5.70M}  \\ 
\rowcolor[gray]{.75}\multicolumn{2}{c}{Big network only}  &  \textbf{25.69\%} $\pm$ 0.30\%  & 7.54M \\ 
\bottomrule
\end{tabular}
\end{table}

\begin{figure}
\centering
\begin{subfigure}{.5\linewidth}
\hspace*{-0.5cm} 
    \begin{tikzpicture}[font=\fontsize{7}{7}\selectfont]
\begin{axis}[
    height=5.5cm,
        width=7cm,
        xmin=10,
        xmax=50,
        ymin=17.5,
        ymax=26.0,
        every axis plot/.append style={thick},
        xlabel=Epoch,
        ylabel=PER,
        x label style={at={(axis description cs:0.5,0.025)},anchor=north},
        y label style={at={(axis description cs:0.1,.5)},anchor=south}
        ]
\addplot[color=red] table[x=x,y=y] {dat/timit_1_0_0.dat};
\addplot [name path=upper,draw=none] table[x=x,y expr=\thisrow{y}+\thisrow{err}] {dat/timit_1_0_0.dat};
\addplot [name path=lower,draw=none] table[x=x,y expr=\thisrow{y}-\thisrow{err}] {dat/timit_1_0_0.dat};
\addplot [fill=red, fill opacity=0.5] fill between[of=upper and lower];



\addplot[color=blue] table[x=x,y=y] {dat/timit_1_1_1.dat};
\addplot [name path=upper,draw=none] table[x=x,y expr=\thisrow{y}+\thisrow{err}] {dat/timit_1_1_1.dat};
\addplot [name path=lower,draw=none] table[x=x,y expr=\thisrow{y}-\thisrow{err}] {dat/timit_1_1_1.dat};
\addplot [fill=blue, fill opacity=0.5] fill between[of=upper and lower];


\legend{,,,surprisal (train) \xmark; surprisal (test) \xmark,,,,surprisal (train) \cmark; surprisal (test) \cmark};

\end{axis}
\end{tikzpicture}
  \caption{Validation PER for TIMIT.}
  \label{fig:valid-timit}
\end{subfigure}%
\begin{subfigure}{.5\linewidth}
    \centering
\hspace*{-0.5cm} 
    \begin{tikzpicture}[font=\fontsize{7}{7}\selectfont]
\begin{axis}[
        height=5.5cm,
        width=7cm,
        xmin=15,
        xmax=46,
        ymin=25.0,
        ymax=33.0,
        every axis plot/.append style={thick},
        xlabel=Epoch,
        ylabel=PER,
        x label style={at={(axis description cs:0.5,0.025)},anchor=north},
        y label style={at={(axis description cs:0.1,.5)},anchor=south}
        ]
\addplot[color=red] table[x=x,y=y] {dat/mini-librispeech_1_0_0.dat};
\addplot [name path=upper,draw=none] table[x=x,y expr=\thisrow{y}+\thisrow{err}] {dat/mini-librispeech_1_0_0.dat};
\addplot [name path=lower,draw=none] table[x=x,y expr=\thisrow{y}-\thisrow{err}] {dat/mini-librispeech_1_0_0.dat};
\addplot [fill=red, fill opacity=0.5] fill between[of=upper and lower];



\addplot[color=blue] table[x=x,y=y] {dat/mini-librispeech_1_1_1.dat};
\addplot [name path=upper,draw=none] table[x=x,y expr=\thisrow{y}+\thisrow{err}] {dat/mini-librispeech_1_1_1.dat};
\addplot [name path=lower,draw=none] table[x=x,y expr=\thisrow{y}-\thisrow{err}] {dat/mini-librispeech_1_1_1.dat};
\addplot [fill=blue, fill opacity=0.5] fill between[of=upper and lower];

\legend{,,,surprisal (train) \xmark; surprisal (test) \xmark,,,,surprisal (train) \cmark; surprisal (test) \cmark};

\end{axis}
\end{tikzpicture}
    \caption{Validation PER for Mini-LibriSpeech.}
    \label{fig:valid-lib}
\end{subfigure}
\caption{Validation PER over the course of training for models with and without surprisal. Similar curves are obtained with a mismatch between train and test (not shown here for clarity of presentation).}
\label{fig:surprisal-curves}
\end{figure}
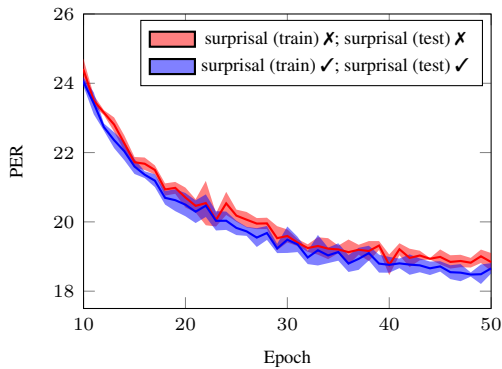
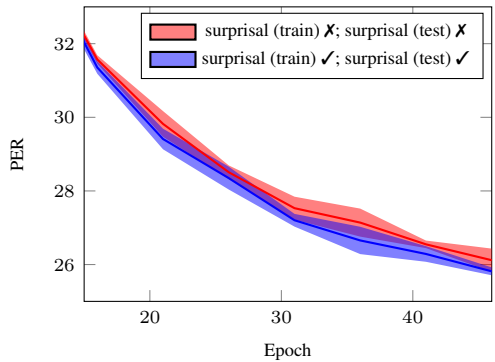

Fig. \ref{fig:surprisal-curves} shows the validation PER of models with and without surprisal-based sampling at test time over the course of training: the models with surprisal-based sampling during training and testing consistently outperform those without over time, despite making slightly less use of the big model (due to the discrepancy between the surprisal levels for the training set and for the test set).

Whether it is crucial that surprisal is used during training is less clear. When surprisal is not used during training, the models that use it during testing do perform slightly better than those that do not for Mini-LibriSpeech (PER of 26.16\% vs. 26.04\%) and TIMIT when not using autoregressive features (22.81\% vs. 22.49\%), but not for TIMIT when using autoregressive features (both 20.52\%). Appendix B further explores the effect of a mismatch between train time and test time.

\section{Conclusion and Future Work}
\label{conclusion}

We have shown that it is possible to use the surprisal of an autoregressive model to determine whether to use a big neural network or a small neural network for processing a stream of inputs, using the big network only for more difficult inputs and thereby reducing overall computation---in one instance reducing FLOP count by 15\% at no cost in accuracy. 
We also find that while a baseline with a learned controller can achieve similar results, our model has  lower variance and is less sensitive to hyperparameters. This suggests that the simple inductive bias of surprisal may make it easier to train much larger conditional models, where rounds of hyperparameter tuning are more expensive.

To return to the analogy with human cognition given in the introduction, a valid criticism of our model is that it does not take into account the expected reward resulting from using more cognitive effort. Given a difficult task, a human decision maker might give more thought to the task if there is a big enough potential reward but might also decide not to waste energy if the reward is small \cite{lieder2018rational}. In this work, we have effectively assumed a uniform potential benefit for using the big network, which turned out to be a reasonable assumption for the tasks we considered; in the future, we hope to compare our technique with a more data-driven evaluation of costs and benefits in a complete reinforcement learning setup. Also, here we have only run experiments in which there is single task to be performed; another interesting setting to study might be multi-tasking, in which an agent must allocate a limited computational budget among various tasks simultaneously \cite{otto2013curse}.

\section*{Broader Impact}

The ethical and societal implications of this work are the same as for any work that deals with the efficiency of neural networks: if neural networks are made more data- and compute-efficient, they become easier to deploy in applications that can have a negative impact on society, like surveillance, though the same can be said for applications with a positive impact. As far as we are aware, there is no additional social dimension to our work that needs to be considered more carefully.

\begin{ack}
Thanks to Ross Otto, Paul Xiong, Tom Bosc, Amir Ardakani, Blake Richards, and Ashutosh Adhikari for helpful feedback and discussions. Thanks to Mirco Ravanelli for help with using TIMIT.
\end{ack}

\medskip

\small

\bibliography{refs}

\begin{thebibliography}{10}
\providecommand{\url}[1]{#1}
\csname url@samestyle\endcsname
\providecommand{\newblock}{\relax}
\providecommand{\bibinfo}[2]{#2}
\providecommand{\BIBentrySTDinterwordspacing}{\spaceskip=0pt\relax}
\providecommand{\BIBentryALTinterwordstretchfactor}{4}
\providecommand{\BIBentryALTinterwordspacing}{\spaceskip=\fontdimen2\font plus
\BIBentryALTinterwordstretchfactor\fontdimen3\font minus
  \fontdimen4\font\relax}
\providecommand{\BIBforeignlanguage}[2]{{%
\expandafter\ifx\csname l@#1\endcsname\relax
\typeout{** WARNING: IEEEtran.bst: No hyphenation pattern has been}%
\typeout{** loaded for the language `#1'. Using the pattern for}%
\typeout{** the default language instead.}%
\else
\language=\csname l@#1\endcsname
\fi
#2}}
\providecommand{\BIBdecl}{\relax}
\BIBdecl

\bibitem{kahneman2011thinking}
D.~Kahneman, \emph{Thinking, fast and slow}.\hskip 1em plus 0.5em minus
  0.4em\relax Macmillan, 2011.

\bibitem{stanovich2000individual}
K.~E. Stanovich and R.~F. West, ``Individual differences in reasoning:
  Implications for the rationality debate?'' \emph{Behavioral and brain
  sciences}, vol.~23, no.~5, pp. 645--665, 2000.

\bibitem{levy2008expectation}
R.~Levy, ``Expectation-based syntactic comprehension,'' \emph{Cognition}, vol.
  106, no.~3, pp. 1126--1177, 2008.

\bibitem{levy2011integrating}
------, ``Integrating surprisal and uncertain-input models in online sentence
  comprehension: formal techniques and empirical results,'' in
  \emph{Proceedings of the 49th Annual Meeting of the Association for
  Computational Linguistics: Human Language Technologies-Volume 1}.\hskip 1em
  plus 0.5em minus 0.4em\relax Association for Computational Linguistics, 2011,
  pp. 1055--1065.

\bibitem{monsalve2012lexical}
I.~F. Monsalve, S.~L. Frank, and G.~Vigliocco, ``Lexical surprisal as a general
  predictor of reading time,'' in \emph{Proceedings of the 13th Conference of
  the European Chapter of the Association for Computational Linguistics}.\hskip
  1em plus 0.5em minus 0.4em\relax Association for Computational Linguistics,
  2012, pp. 398--408.

\bibitem{amodei31ai}
D.~Amodei and D.~Hernandez, ``{AI and Compute},'' \emph{OpenAI blog
  https://blog.openai. com/ai-and-compute}, vol.~31.

\bibitem{sanh2019distilbert}
V.~Sanh, L.~Debut, J.~Chaumond, and T.~Wolf, ``{DistilBERT, a distilled version
  of BERT: smaller, faster, cheaper and lighter},'' \emph{NeurIPS Workshop on
  Energy Efficient Machine Learning and Cognitive Computing}, 2019.

\bibitem{strubell2019energy}
E.~Strubell, A.~Ganesh, and A.~McCallum, ``{Energy and policy considerations
  for deep learning in NLP},'' \emph{57th Annual Meeting of the Association for
  Computational Linguistics (ACL)}, 2019.

\bibitem{sharir2020cost}
O.~Sharir, B.~Peleg, and Y.~Shoham, ``{The Cost of Training NLP Models: A
  Concise Overview},'' \emph{arXiv preprint arXiv:2004.08900}, 2020.

\bibitem{chapelle2009semi}
\BIBentryALTinterwordspacing
O.~Chapelle, B.~Schölkopf, and A.~Zien, Eds., \emph{Semi-Supervised
  Learning}.\hskip 1em plus 0.5em minus 0.4em\relax Cambridge, MA: MIT Press,
  2006. [Online]. Available: \url{http://www.olivier.chapelle.cc/ssl-book/}
\BIBentrySTDinterwordspacing

\bibitem{bengio2013estimating}
Y.~Bengio, N.~L{\'e}onard, and A.~Courville, ``Estimating or propagating
  gradients through stochastic neurons for conditional computation,''
  \emph{arXiv preprint arXiv:1308.3432}, 2013.

\bibitem{davis2013low}
A.~Davis and I.~Arel, ``Low-rank approximations for conditional feedforward
  computation in deep neural networks,'' \emph{arXiv preprint arXiv:1312.4461},
  2013.

\bibitem{teerapittayanon2016branchynet}
S.~Teerapittayanon, B.~McDanel, and H.-T. Kung, ``Branchynet: Fast inference
  via early exiting from deep neural networks,'' in \emph{2016 23rd
  International Conference on Pattern Recognition (ICPR)}.\hskip 1em plus 0.5em
  minus 0.4em\relax IEEE, 2016, pp. 2464--2469.

\bibitem{tan2016towards}
S.~Tan and K.~C. Sim, ``Towards implicit complexity control using
  variable-depth deep neural networks for automatic speech recognition,'' in
  \emph{2016 IEEE International Conference on Acoustics, Speech and Signal
  Processing (ICASSP)}.\hskip 1em plus 0.5em minus 0.4em\relax IEEE, 2016, pp.
  5965--5969.

\bibitem{Xin2020deebert}
J.~Xin, R.~Tang, J.~Lee, Y.~Yu, and J.~Lin, ``{DeeBERT: Dynamic Early Exiting
  for Accelerating BERT Inference},'' \emph{ACL}, 2020.

\bibitem{stamoulis2018designing}
D.~Stamoulis, T.-W. Chin, A.~K. Prakash, H.~Fang, S.~Sajja, M.~Bognar, and
  D.~Marculescu, ``Designing adaptive neural networks for energy-constrained
  image classification,'' in \emph{Proceedings of the International Conference
  on Computer-Aided Design}, 2018, pp. 1--8.

\bibitem{scardapane2020add}
S.~Scardapane, M.~Scarpiniti, E.~Baccarelli, and A.~Uncini, ``Why should we add
  early exits to neural networks?'' 2020.

\bibitem{park2015big}
E.~Park, D.~Kim, S.~Kim, Y.-D. Kim, G.~Kim, S.~Yoon, and S.~Yoo, ``Big/little
  deep neural network for ultra low power inference,'' in \emph{2015
  International Conference on Hardware/Software Codesign and System Synthesis
  (CODES+ ISSS)}.\hskip 1em plus 0.5em minus 0.4em\relax IEEE, 2015, pp.
  124--132.

\bibitem{gruenstein2017cascade}
A.~Gruenstein, R.~Alvarez, C.~Thornton, and M.~Ghodrat, ``A cascade
  architecture for keyword spotting on mobile devices,'' \emph{NeurIPS -
  Workshop on Machine Learning on the Phone and other Consumer Devices}, 2017.

\bibitem{tann2018flexible}
H.~Tann, S.~Hashemi, and S.~Reda, ``Flexible deep neural network processing,''
  \emph{arXiv preprint arXiv:1801.07353}, 2018.

\bibitem{pagliari2018dynamic}
D.~J. Pagliari, E.~Macii, and M.~Poncino, ``Dynamic bit-width reconfiguration
  for energy-efficient deep learning hardware,'' in \emph{Proceedings of the
  International Symposium on Low Power Electronics and Design}, 2018, pp. 1--6.

\bibitem{venkataramani2015scalable}
S.~Venkataramani, A.~Raghunathan, J.~Liu, and M.~Shoaib, ``Scalable-effort
  classifiers for energy-efficient machine learning,'' in \emph{Proceedings of
  the 52nd Annual Design Automation Conference}, 2015, pp. 1--6.

\bibitem{jacobs1991adaptive}
R.~A. Jacobs, M.~I. Jordan, S.~J. Nowlan, and G.~E. Hinton, ``Adaptive mixtures
  of local experts,'' \emph{Neural computation}, vol.~3, no.~1, pp. 79--87,
  1991.

\bibitem{masoudnia2014mixture}
S.~Masoudnia and R.~Ebrahimpour, ``Mixture of experts: a literature survey,''
  \emph{Artificial Intelligence Review}, vol.~42, no.~2, pp. 275--293, 2014.

\bibitem{bengio2015conditional}
E.~Bengio, P.-L. Bacon, J.~Pineau, and D.~Precup, ``Conditional computation in
  neural networks for faster models,'' \emph{ICLR}, 2016.

\bibitem{leonard2015distributed}
N.~L{\'e}onard, ``Distributed conditional computation,'' 2015.

\bibitem{yu2017learning}
A.~W. Yu, H.~Lee, and Q.~V. Le, ``Learning to skim text,'' \emph{arXiv preprint
  arXiv:1704.06877}, 2017.

\bibitem{odena2017changing}
A.~Odena, D.~Lawson, and C.~Olah, ``Changing model behavior at test-time using
  reinforcement learning,'' \emph{ICLR Workshop}, 2017.

\bibitem{liu2018dynamic}
L.~Liu and J.~Deng, ``Dynamic deep neural networks: Optimizing
  accuracy-efficiency trade-offs by selective execution,'' in
  \emph{Thirty-Second AAAI Conference on Artificial Intelligence}, 2018.

\bibitem{dean2019deep}
J.~Dean, ``The deep learning revolution and its implications for computer
  architecture and chip design,'' in \emph{2020 IEEE International Solid-State
  Circuits Conference-(ISSCC)}.\hskip 1em plus 0.5em minus 0.4em\relax IEEE,
  2020, pp. 8--14.

\bibitem{eigen2013learning}
D.~Eigen, M.~Ranzato, and I.~Sutskever, ``Learning factored representations in
  a deep mixture of experts,'' \emph{ICLR Workshop}, 2014.

\bibitem{cho2014exponentially}
K.~Cho and Y.~Bengio, ``Exponentially increasing the capacity-to-computation
  ratio for conditional computation in deep learning,'' \emph{arXiv preprint
  arXiv:1406.7362}, 2014.

\bibitem{rosenbaum2017routing}
C.~Rosenbaum, T.~Klinger, and M.~Riemer, ``Routing networks: Adaptive selection
  of non-linear functions for multi-task learning,'' \emph{ICLR}, 2018.

\bibitem{tanno2018adaptive}
R.~Tanno, K.~Arulkumaran, D.~C. Alexander, A.~Criminisi, and A.~Nori,
  ``Adaptive neural trees,'' \emph{ICML}, 2019.

\bibitem{graves2016adaptive}
A.~Graves, ``Adaptive computation time for recurrent neural networks,''
  \emph{arXiv preprint arXiv:1603.08983}, 2016.

\bibitem{figurnov2017spatially}
M.~Figurnov, M.~D. Collins, Y.~Zhu, L.~Zhang, J.~Huang, D.~Vetrov, and
  R.~Salakhutdinov, ``Spatially adaptive computation time for residual
  networks,'' in \emph{Proceedings of the IEEE Conference on Computer Vision
  and Pattern Recognition}, 2017, pp. 1039--1048.

\bibitem{shazeer2017outrageously}
N.~Shazeer, A.~Mirhoseini, K.~Maziarz, A.~Davis, Q.~Le, G.~Hinton, and J.~Dean,
  ``Outrageously large neural networks: The sparsely-gated mixture-of-experts
  layer,'' \emph{ICLR}, 2017.

\bibitem{campos2017skip}
V.~Campos, B.~Jou, X.~Gir{\'o}-i Nieto, J.~Torres, and S.-F. Chang, ``Skip
  {RNN}: Learning to skip state updates in recurrent neural networks,''
  \emph{NeurIPS Time Series Workshop}, 2017.

\bibitem{jernite2016variable}
Y.~Jernite, E.~Grave, A.~Joulin, and T.~Mikolov, ``Variable computation in
  recurrent neural networks,'' \emph{ICLR}, 2017.

\bibitem{bolukbasi2017adaptive}
T.~Bolukbasi, J.~Wang, O.~Dekel, and V.~Saligrama, ``Adaptive neural networks
  for efficient inference,'' in \emph{Proceedings of the 34th International
  Conference on Machine Learning-Volume 70}.\hskip 1em plus 0.5em minus
  0.4em\relax JMLR. org, 2017, pp. 527--536.

\bibitem{dehghani2018universal}
M.~Dehghani, S.~Gouws, O.~Vinyals, J.~Uszkoreit, and {\L}.~Kaiser, ``Universal
  transformers,'' \emph{ICLR}, 2019.

\bibitem{bapna2020controlling}
A.~Bapna, N.~Arivazhagan, and O.~Firat, ``Controlling computation versus
  quality for neural sequence models,'' \emph{arXiv preprint arXiv:2002.07106},
  2020.

\bibitem{NIPS2019_9611}
J.~Ren, P.~J. Liu, E.~Fertig, J.~Snoek, R.~Poplin, M.~Depristo, J.~Dillon, and
  B.~Lakshminarayanan, ``Likelihood ratios for out-of-distribution detection,''
  in \emph{Advances in Neural Information Processing Systems 32}, 2019.

\bibitem{mielke2019kind}
S.~J. Mielke, R.~Cotterell, K.~Gorman, B.~Roark, and J.~Eisner, ``What kind of
  language is hard to language-model?'' \emph{ACL}, 2019.

\bibitem{he2019pun}
H.~He, N.~Peng, and P.~Liang, ``Pun generation with surprise,'' \emph{NAACL},
  2019.

\bibitem{rocki2016surprisal}
K.~M. Rocki, ``Surprisal-driven feedback in recurrent networks,'' \emph{arXiv
  preprint arXiv:1608.06027}, 2016.

\bibitem{rocki2016zoneout}
K.~Rocki, T.~Kornuta, and T.~Maharaj, ``Surprisal-driven zoneout,''
  \emph{NeurIPS - Continual Learning and Deep Networks Workshop}, 2016.

\bibitem{alpay2019preserving}
T.~Alpay, F.~Abawi, and S.~Wermter, ``Preserving activations in recurrent
  neural networks based on surprisal,'' \emph{Neurocomputing}, vol. 342, pp.
  75--82, 2019.

\bibitem{dai2015semi}
A.~M. Dai and Q.~V. Le, ``Semi-supervised sequence learning,'' in
  \emph{Advances in neural information processing systems}, 2015, pp.
  3079--3087.

\bibitem{lotter2016deep}
W.~Lotter, G.~Kreiman, and D.~Cox, ``Deep predictive coding networks for video
  prediction and unsupervised learning,'' \emph{ICLR}, 2017.

\bibitem{radford2017learning}
A.~Radford, R.~Jozefowicz, and I.~Sutskever, ``Learning to generate reviews and
  discovering sentiment,'' \emph{arXiv preprint arXiv:1704.01444}, 2017.

\bibitem{howard2018universal}
J.~Howard and S.~Ruder, ``Universal language model fine-tuning for text
  classification,'' \emph{ACL}, 2018.

\bibitem{ha2018world}
D.~Ha and J.~Schmidhuber, ``Recurrent world models facilitate policy
  evolution,'' \emph{NeurIPS}, 2018.

\bibitem{chung2019unsupervised}
Y.-A. Chung, W.-N. Hsu, H.~Tang, and J.~Glass, ``An unsupervised autoregressive
  model for speech representation learning,'' \emph{Interspeech}, 2019.

\bibitem{yang2019xlnet}
Z.~Yang, Z.~Dai, Y.~Yang, J.~Carbonell, R.~R. Salakhutdinov, and Q.~V. Le,
  ``Xlnet: Generalized autoregressive pretraining for language understanding,''
  in \emph{Advances in neural information processing systems}, 2019, pp.
  5754--5764.

\bibitem{srivastava2015unsupervised}
N.~Srivastava, E.~Mansimov, and R.~Salakhudinov, ``{Unsupervised learning of
  video representations using LSTMs},'' in \emph{International Conference on
  Machine Learning}, 2015, pp. 843--852.

\bibitem{radford2019language}
A.~Radford, J.~Wu, R.~Child, D.~Luan, D.~Amodei, and I.~Sutskever, ``Language
  models are unsupervised multitask learners,'' \emph{OpenAI Blog}, vol.~1,
  no.~8, p.~9, 2019.

\bibitem{rissanen1976generalized}
J.~J. Rissanen, ``{Generalized Kraft inequality and arithmetic coding},''
  \emph{IBM Journal of Research and Development}, vol.~20, no.~3, pp. 198--203,
  1976.

\bibitem{witten1987arithmetic}
I.~H. Witten, R.~M. Neal, and J.~G. Cleary, ``Arithmetic coding for data
  compression,'' \emph{Communications of the ACM}, vol.~30, no.~6, pp.
  520--540, 1987.

\bibitem{frantz1982design}
G.~A. Frantz and R.~H. Wiggins, ``{Design case history: Speak \& Spell learns
  to talk},'' \emph{IEEE Spectrum}, vol.~19, no.~2, pp. 45--49, 1982.

\bibitem{o1988linear}
D.~O'Shaughnessy, ``Linear predictive coding,'' \emph{IEEE Potentials}, vol.~7,
  no.~1, pp. 29--32, 1988.

\bibitem{oord2018representation}
A.~v.~d. Oord, Y.~Li, and O.~Vinyals, ``Representation learning with
  contrastive predictive coding,'' \emph{arXiv preprint arXiv:1807.03748},
  2018.

\bibitem{huang2011predictive}
Y.~Huang and R.~P. Rao, ``Predictive coding,'' \emph{Wiley Interdisciplinary
  Reviews: Cognitive Science}, vol.~2, no.~5, pp. 580--593, 2011.

\bibitem{clark2013whatever}
A.~Clark, ``{Whatever next? Predictive brains, situated agents, and the future
  of cognitive science},'' \emph{Behavioral and Brain Sciences}, vol.~36,
  no.~3, pp. 181--204, 2013.

\bibitem{schmidhuber1991neural}
J.~Schmidhuber, ``Neural sequence chunkers,'' 1991.

\bibitem{aljundi2017expert}
R.~Aljundi, P.~Chakravarty, and T.~Tuytelaars, ``Expert gate: Lifelong learning
  with a network of experts,'' in \emph{Proceedings of the IEEE Conference on
  Computer Vision and Pattern Recognition}, 2017, pp. 3366--3375.

\bibitem{sheikhnezhad2019novel}
F.~S.~Fard and T.~Trappenberg, ``A novel model for arbitration between planning
  and habitual control systems,'' \emph{Frontiers in Neurorobotics}, vol.~13,
  p.~52, 2019.

\bibitem{goyalvariational}
A.~Goyal, Y.~Bengio, and M.~B.~S. Levine, ``The variational bandwidth
  bottleneck: Stochastic evaluation on an information budget,'' \emph{ICLR},
  2020.

\bibitem{elman1990finding}
J.~L. Elman, ``Finding structure in time,'' \emph{Cognitive science}, vol.~14,
  no.~2, pp. 179--211, 1990.

\bibitem{bengio2003neural}
Y.~Bengio, R.~Ducharme, P.~Vincent, and C.~Jauvin, ``A neural probabilistic
  language model,'' \emph{Journal of machine learning research}, vol.~3, no.
  Feb, pp. 1137--1155, 2003.

\bibitem{larochelle2011neural}
H.~Larochelle and I.~Murray, ``The neural autoregressive distribution
  estimator,'' in \emph{Proceedings of the Fourteenth International Conference
  on Artificial Intelligence and Statistics}, 2011, pp. 29--37.

\bibitem{gehring2017convolutional}
J.~Gehring, M.~Auli, D.~Grangier, D.~Yarats, and Y.~N. Dauphin, ``Convolutional
  sequence to sequence learning,'' in \emph{Proceedings of the 34th
  International Conference on Machine Learning-Volume 70}.\hskip 1em plus 0.5em
  minus 0.4em\relax JMLR. org, 2017, pp. 1243--1252.

\bibitem{vaswani2017attention}
A.~Vaswani, N.~Shazeer, N.~Parmar, J.~Uszkoreit, L.~Jones, A.~N. Gomez,
  {\L}.~Kaiser, and I.~Polosukhin, ``Attention is all you need,'' in
  \emph{Advances in neural information processing systems}, 2017, pp.
  5998--6008.

\bibitem{peters2019tune}
M.~Peters, S.~Ruder, and N.~A. Smith, ``{To tune or not to tune? Adapting
  pretrained representations to diverse tasks},'' \emph{4th Workshop on
  Representation Learning for NLP}, 2019.

\bibitem{chen2018neural}
T.~Q. Chen, Y.~Rubanova, J.~Bettencourt, and D.~K. Duvenaud, ``Neural ordinary
  differential equations,'' in \emph{Advances in neural information processing
  systems}, 2018, pp. 6571--6583.

\bibitem{paszke2019pytorch}
A.~Paszke, S.~Gross, F.~Massa, A.~Lerer, J.~Bradbury, G.~Chanan, T.~Killeen,
  Z.~Lin, N.~Gimelshein, L.~Antiga \emph{et~al.}, ``{PyTorch: An imperative
  style, high-performance deep learning library},'' in \emph{{Advances in
  Neural Information Processing Systems}}, 2019, pp. 8024--8035.

\bibitem{garofolo1993darpa}
J.~S. Garofolo, L.~F. Lamel, W.~M. Fisher, J.~G. Fiscus, and D.~S. Pallett,
  ``{DARPA TIMIT acoustic-phonetic continous speech corpus CD-ROM. NIST speech
  disc 1-1.1},'' \emph{NASA STI/Recon technical report n}, vol.~93, 1993.

\bibitem{panayotov2015librispeech}
V.~Panayotov, G.~Chen, D.~Povey, and S.~Khudanpur, ``{LibriSpeech: an ASR
  corpus based on public domain audio books},'' in \emph{2015 IEEE
  International Conference on Acoustics, Speech and Signal Processing
  (ICASSP)}.\hskip 1em plus 0.5em minus 0.4em\relax IEEE, 2015, pp. 5206--5210.

\bibitem{mcauliffe2017montreal}
M.~McAuliffe, M.~Socolof, S.~Mihuc, M.~Wagner, and M.~Sonderegger, ``{Montreal
  Forced Aligner: Trainable Text-Speech Alignment Using Kaldi.}'' in
  \emph{Interspeech}, 2017, pp. 498--502.

\bibitem{chung2014empirical}
J.~Chung, C.~Gulcehre, K.~Cho, and Y.~Bengio, ``Empirical evaluation of gated
  recurrent neural networks on sequence modeling,'' \emph{NeurIPS 2014 Deep
  Learning and Representation Learning Workshop}, 2014.

\bibitem{zaremba2014recurrent}
W.~Zaremba, I.~Sutskever, and O.~Vinyals, ``Recurrent neural network
  regularization,'' \emph{arXiv preprint arXiv:1409.2329}, 2014.

\bibitem{sak2015fast}
H.~Sak, A.~Senior, K.~Rao, and F.~Beaufays, ``Fast and accurate recurrent
  neural network acoustic models for speech recognition,'' \emph{Interspeech},
  2015.

\bibitem{graves2006connectionist}
A.~Graves, S.~Fern{\'a}ndez, F.~Gomez, and J.~Schmidhuber, ``Connectionist
  temporal classification: labelling unsegmented sequence data with recurrent
  neural networks,'' in \emph{Proceedings of the 23rd international conference
  on Machine learning}, 2006, pp. 369--376.

\bibitem{hannun_2017}
\BIBentryALTinterwordspacing
A.~Hannun, ``Training sequence models with attention,'' 2017. [Online].
  Available: \url{https://awni.github.io/train-sequence-models/}
\BIBentrySTDinterwordspacing

\bibitem{brown2020language}
T.~B. Brown, B.~Mann, N.~Ryder, M.~Subbiah, J.~Kaplan, P.~Dhariwal,
  A.~Neelakantan, P.~Shyam, G.~Sastry, A.~Askell, S.~Agarwal, A.~Herbert-Voss,
  G.~Krueger, T.~Henighan, R.~Child, A.~Ramesh, D.~M. Ziegler, J.~Wu,
  C.~Winter, C.~Hesse, M.~Chen, E.~Sigler, M.~Litwin, S.~Gray, B.~Chess,
  J.~Clark, C.~Berner, S.~McCandlish, A.~Radford, I.~Sutskever, and D.~Amodei,
  ``Language models are few-shot learners,'' 2020.

\bibitem{lieder2018rational}
F.~Lieder, A.~Shenhav, S.~Musslick, and T.~L. Griffiths, ``Rational
  metareasoning and the plasticity of cognitive control,'' \emph{PLoS
  computational biology}, vol.~14, no.~4, p. e1006043, 2018.

\bibitem{otto2013curse}
A.~R. Otto, S.~J. Gershman, A.~B. Markman, and N.~D. Daw, ``The curse of
  planning: dissecting multiple reinforcement-learning systems by taxing the
  central executive,'' \emph{Psychological science}, vol.~24, no.~5, pp.
  751--761, 2013.

\end{thebibliography}
\bibliographystyle{IEEEtran}

\clearpage
\normalsize
\section*{Appendix A: Effect of varying hyperparameters}
We re-ran the ablation experiment on TIMIT using three other neural architectures different from the one used in our main experiments, listed here in more concise PyTorch-like notation. For these models, the controller trained to have mean 0.65 for $p_{big}$, instead of 0.5. The results show that using surprisal for effort allocation at test time improves PER across different hyperparameter settings, and that the pre-net, post-net, and recurrent layers are not essential for the idea to work. Using different hyperparameters for the autoregressive model would also be worthwhile to explore, but as training it is time-consuming, we have restricted the autoregressive model to a single configuration.

\begin{wraptable}{r}{10cm}
\caption{Model 1 results for TIMIT.}
\label{model-1}
\begin{tabular}{cccc} 
\toprule
\thead{Surprisal-based\\ during training?} & \thead{Surprisal-based\\ during testing?} & \thead{PER} & \thead{Avg. FLOPs\\per input}\\
\midrule
\xmark & \xmark     & 26.94\% $\pm$ 0.48\%  &  6.52M     \\ %
\rowcolor[gray]{.75}\xmark  & \cmark  &  26.89\% $\pm$ 0.20\%   & 6.49M  \\ %
\cmark & \xmark  &  28.86\% $\pm$ 0.31\%  &  6.52M     \\%
\cmark & \cmark  & 27.51\% $\pm$ 0.39\%  & 6.49M   \\ 
\midrule
\rowcolor[gray]{.75}\multicolumn{2}{c}{Small network only}   & 27.72\% $\pm$ 0.53\% &  \textbf{5.96M}  \\ 
\rowcolor[gray]{.75}\multicolumn{2}{c}{Big network only}  &  \textbf{26.75\%} $\pm$ 0.13\%  & 7.07M \\ 
\bottomrule
\end{tabular}
\end{wraptable}

\textbf{Model 1:}

\textit{Pre-net:} \\
\texttt{Conv1D(length 11, 512 filters)\\
LeakyReLU(0.125)}

\textit{Small network:}\\
\texttt{Linear(512, 40)\\
LogSoftmax()}

\textit{Big network:}\\
\texttt{Linear(512, 2048)\\
LeakyReLU(0.125)\\
Linear(2048, 40)\\
LogSoftmax()}

\textit{Post-net:} (none)

\begin{wraptable}{r}{9cm}
\caption{Model 2 results for TIMIT.}
\label{model-2}
\begin{tabular}{cccc} 
\toprule
\thead{Surprisal-based\\ during training?} & \thead{Surprisal-based\\ during testing?} & \thead{PER} & \thead{Avg. FLOPs\\per input}\\
\midrule
\xmark & \xmark     &  26.76\% $\pm$ 0.11\% &  6.52M     \\ %
\rowcolor[gray]{.75}\xmark  & \cmark  & 26.30\% $\pm$ 0.36\% & 6.49M  \\ %
\cmark & \xmark  &  28.16\% $\pm$ 0.27\%  & 6.52M \\%
\cmark & \cmark  & 26.54\% $\pm$ 0.32\%  & 6.49M  \\ %
\midrule
\rowcolor[gray]{.75}\multicolumn{2}{c}{Small network only}   &  27.60\% $\pm$ 0.60\% &  \textbf{5.96M}  \\ %
\rowcolor[gray]{.75}\multicolumn{2}{c}{Big network only}  & \textbf{25.91\%} $\pm$ 0.41\%   & 7.07M \\ %
\bottomrule
\end{tabular}
\end{wraptable}

\vspace{0.25cm}
\textbf{Model 2:}

Model 2 is the same as Model 1, except we add \texttt{Dropout(0.5)} to the end of the pre-net. (We found that the variants of Model 1 trained with surprisal easily overfit, which we believe happens because it is easier for the big and small networks to model either only surprising inputs or only unsurprising inputs separately than a mixture of both types.)

\begin{wraptable}{r}{10cm}
\caption{Model 3 results for TIMIT.}
\label{model-3}
\begin{tabular}{cccc}
\toprule
\thead{Surprisal-based\\ during training?} & \thead{Surprisal-based\\ during testing?} & \thead{PER} & \thead{Avg. FLOPs\\per input}\\
\midrule
\xmark & \xmark     &  32.30\% $\pm$ 0.66\%  &   4.62M    \\ %
\xmark  & \cmark  &  31.14\% $\pm$ 0.54\%  & 4.54M   \\ %
\cmark & \xmark  &  34.07\% $\pm$ 0.58\%   &  4.62M    \\%
\rowcolor[gray]{.75}\cmark & \cmark  &  30.37\% $\pm$ 0.31\%  &  4.54M   \\ %
\midrule
\rowcolor[gray]{.75}\multicolumn{2}{c}{Small network only}   & 35.47\% $\pm$ 0.68\%  &   \textbf{3.28M}  \\ 
\rowcolor[gray]{.75}\multicolumn{2}{c}{Big network only}  &  \textbf{28.04\%}  $\pm$ 0.55\%  & 5.96M \\ 
\bottomrule
\end{tabular}
\end{wraptable}

\vspace{0.25cm}
\textbf{Model 3:}

\textit{Pre-net:} (none)

\textit{Small network:}\\
\texttt{Dropout(0.5),\\
Conv1D(length 11, 40 filters)\\
LogSoftmax()}

\textit{Big network:}\\
\texttt{Dropout(0.5),\\
Conv1D(length 11, 512 filters)\\
LeakyReLU(0.125)\\
Linear(512, 40)\\
LogSoftmax()}

\textit{Post-net:} (none)

\clearpage
\section*{Appendix B: Effect of varying controller bias}

To investigate the effect of a mismatch between train time and test time further, we ran an experiment with Mini-LibriSpeech in which the bias parameter of the controller of models trained with surprisal is gradually increased from (average bias value learned when training the controller) to (that value $+$ 4) in uniform increments. This increases the probability of using the big network from what it was during training. 

\begin{figure}[h]
    \centering
\hspace*{-0.5cm} 
    \begin{tikzpicture}[font=\fontsize{7}{7}\selectfont]
\begin{axis}[
    legend style={at={(0.7,0.175)},anchor=north},
        height=5.5cm,
        width=7cm,
        xmin=-1.25,
        xmax=3,
        ymin=25.6,
        ymax=26,
        every axis plot/.append style={thick},
        xlabel=Controller bias,
        ylabel=PER,
        x label style={at={(axis description cs:0.5,0.025)},anchor=north},
        y label style={at={(axis description cs:0.1,.5)},anchor=south}
        ]
\addplot[color=blue] table[x=x,y=y] {dat/sweep.dat};

\addplot [name path=upper,draw=none] table[x=x,y expr=\thisrow{y}+\thisrow{err}] {dat/sweep.dat};
\addplot [name path=lower,draw=none] table[x=x,y expr=\thisrow{y}-\thisrow{err}] {dat/sweep.dat};
\addplot [fill=blue, fill opacity=0.5] fill between[of=upper and lower];

\addplot[color=black, dashed] table[x=x,y=y] {dat/big_only.dat};

\legend{,,,,Big network only};

\end{axis}
\end{tikzpicture}
    \caption{Test PER for Mini-LibriSpeech when increasing the controller bias.} 
    \label{fig:sweep}
\end{figure}
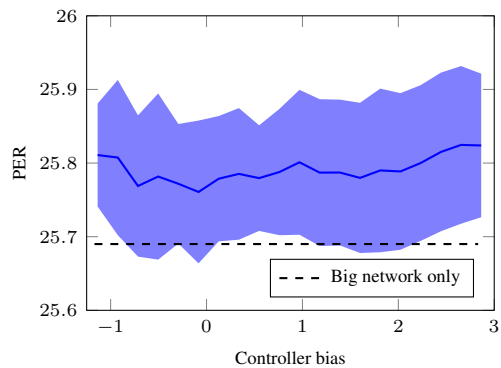

The resulting sweep of mean test PERs with standard deviations is shown in Figure \ref{fig:sweep}. Using the big network more does decrease PER up to a point, but when the bias is increased to the point where the big network is always sampled, the conditional model does not attain the same PER as the model that only ever uses the big network during training and testing. In fact, PER begins to increase, which may be because the big network is exposed to a distribution of more unsurprising inputs than it was shown during training.

In general, though, always using the big network may not be expected to yield the best performance; using both the big and small networks at different times may have an ensemble effect, which may explain how the conditional models in our main experiments were able to slightly outperform the model using only the big network for TIMIT.

\clearpage
\section*{Appendix C: Effect of deterministic execution}

In our main experiments, the big network is selected stochastically with probability $p_{big}$, rather than deterministically when $p_{big}$ is greater than 0.5. This makes the comparison with randomized controllers in the ablation study more appropriate and removes the possibility of certain unlucky edge cases---for example, if $p_{big}$ is just barely less than 0.5 for the entirety of an input sequence, the big network would never be sampled in deterministic execution.

However, it is sometimes desirable for the execution of a model to be deterministic, e.g.~to make software testing easier and to obviate the need for random number generation when implementing the model in on a resource-limited device. We therefore also report the test results when the big network is selected deterministically in Tables \ref{det-ms} and \ref{det-timit}. It appears that when training stochastically, deterministic execution at test time causes a small increase in PER, though this is not reflected in the test loss. Also, it appears crucial to train stochastically, as otherwise overfitting occurs more easily, possibly because the big and small networks always see the same input observations.

\begin{table}[h]
\caption{Results comparing deterministic and stochastic execution for Mini-LibriSpeech.}
\label{det-ms}
\begin{tabular}{cccccc}
\toprule
\thead{Stochastic\\ during training?} & \thead{Stochastic\\ during testing?} & \thead{Train loss} & \thead{Test loss} & \thead{Test PER} & \thead{Avg. FLOPs\\per input}\\
\midrule
\xmark & \xmark     & 69.96 $\pm$ 0.41 & 62.80 $\pm$ 0.76 &  26.45\% $\pm$ 0.23\% &  6.43M       \\ %
\xmark  & \cmark  & 69.72  $\pm$ 0.43 &  63.69 $\pm$ 0.36 & 26.48\% $\pm$ 0.14\% &  6.43M    \\ %
\cmark & \xmark  &  73.76  $\pm$ 0.31  & 62.15 $\pm$ 0.60 & 25.98\% $\pm$ 0.26\% &   6.43M    \\%
\cmark & \cmark  &  74.30 $\pm$ 0.52  & 62.18 $\pm$ 0.64 & \textbf{25.80\%} $\pm$ 0.10\%  & 6.43M   \\  
\bottomrule
\end{tabular}
\end{table}

\begin{table}[h]
\caption{Results comparing deterministic and stochastic execution for TIMIT.}
\label{det-timit}
\begin{tabular}{cccccc}
\toprule
\thead{Stochastic\\ during training?} & \thead{Stochastic\\ during testing?} & \thead{Train loss} & \thead{Test loss} & \thead{Test PER} & \thead{Avg. FLOPs\\per input}\\
\midrule
\xmark & \xmark     & 16.19 $\pm$ 0.14  &  24.61 $\pm$ 0.52 &  21.00\% $\pm$ 0.19\%  &  6.41M       \\ %
\xmark  & \cmark  & 16.21 $\pm$ 0.08 & 25.11 $\pm$ 0.24  & 20.81\% $\pm$ 0.24\% &  6.41M    \\ %
\cmark & \xmark  &  17.58  $\pm$ 0.20 & 23.88 $\pm$ 0.31 & 20.08\% $\pm$ 0.22\% &   6.41M    \\%
\cmark & \cmark  &  17.55 $\pm$ 0.26  &  24.04 $\pm$ 0.22  &  \textbf{20.00\%} $\pm$ 0.15\%   & 6.41M   \\  
\bottomrule
\end{tabular}
\end{table}

(It is possible to train a model either stochastically or deterministically, and then test it both stochastically and deterministically, thus avoiding redundant training runs. The train loss is only different between the first and second rows and between the third and fourth rows because it did not occur to us at the time of this experiment to implement this optimization.)

\clearpage
\section*{Appendix D: More detail on the learned controller baseline}

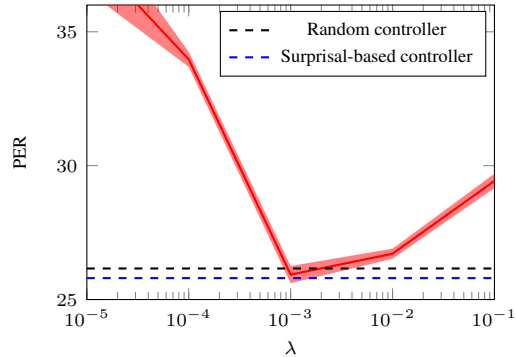
\begin{figure}[h]
    \centering
\hspace*{-0.5cm} 
    \begin{tikzpicture}[font=\fontsize{7}{7}\selectfont]
\begin{axis}[
        height=5.5cm,
        width=7cm,
        xmode = log,
        xmin=0.00001,
        xmax=0.1,
        ymin=25,
        ymax=36,
        every axis plot/.append style={thick},
        xlabel=$\lambda$,
        ylabel=PER,
        x label style={at={(axis description cs:0.5,0.025)},anchor=north},
        y label style={at={(axis description cs:0.1,.5)},anchor=south}
        ]
\addplot[color=red] table[x=x,y=y] {dat/learnedcontroller.dat};

\addplot [name path=upper,draw=none] table[x=x,y expr=\thisrow{y}+\thisrow{err}] {dat/learnedcontroller.dat};
\addplot [name path=lower,draw=none] table[x=x,y expr=\thisrow{y}-\thisrow{err}] {dat/learnedcontroller.dat};
\addplot [fill=red, fill opacity=0.5] fill between[of=upper and lower];

\addplot[color=black, dashed] table[x=x,y=y] {dat/randomcontroller.dat};

\addplot[color=blue, dashed] table[x=x,y=y] {dat/surprisalcontroller.dat};

\legend{,,,,Random controller,Surprisal-based controller};

\end{axis}
\end{tikzpicture}
    \caption{Test PER for Mini-LibriSpeech for models with a learned controller as a function of $\lambda$.}
    \label{fig:learnedcontroller}
\end{figure}

When comparing the performance of surprisal-based controllers with learned controllers in Section \ref{sec:compare}, we noted that the learned controllers' performance is sensitive to the hyperparameter $\lambda$ used in the loss function. This observation is illustrated for Mini-LibriSpeech in Figure \ref{fig:learnedcontroller}. 

Of course, this result should be taken with a grain of salt: the space of possible models for implementing the learned controller is vast, and there may be others that are more robust. Still, the fact that a gating network using the straight-through estimator---the simplest and best-performing approach to learned conditional computation considered in \cite{bengio2013estimating}---behaves this way suggests that other learned conditional computation methods, such as those based on reinforcement learning, may encounter similar difficulties.

\end{document}